\documentclass[letterpaper, 10 pt, conferenceStab]{ieeeconf}
\IEEEoverridecommandlockouts
\usepackage{cite}
\usepackage{amsmath,amssymb,mathrsfs,amsfonts}
\usepackage{bbm}
\usepackage{algorithmic}
\usepackage{graphicx}
\usepackage{subcaption}
\usepackage{color}
\usepackage[font=small]{caption}

\newtheorem{theorem}{Theorem}

\newtheorem{remark}{Remark}

\begin{document}

\title{\bf Stable Real-Time Feedback Control of a Pneumatic Soft Robot}
    
\author{Sean Even*, Tongjia Zheng*, Hai Lin, Yasemin Ozkan-Aydin,\textit{ Member, IEEE}
\thanks{*These authors contributed equally. }
\thanks{$^{}$This work was not supported by any organization. All the authors are with the Department of Electrical Engineering, University of Notre Dame, Notre Dame, IN 46556 USA
        {\tt\small seven,tzheng1,hlin1,yozkanay@nd.edu}}
}

\maketitle
\thispagestyle{empty}
\pagestyle{empty}

\begin{abstract}
Soft actuators offer compliant and safe interaction with an unstructured environment  compared to their rigid counterparts. However, control of these systems is often challenging because they are inherently under-actuated, have infinite degrees of freedom (DoF), and their mechanical properties can change by unknown external loads. 
Existing works mainly relied on discretization and reduction, suffering from either low accuracy or high computational cost for real-time control purposes. Recently, we presented an infinite-dimensional feedback controller for soft manipulators modeled by partial differential equations (PDEs) based on the Cosserat rod theory. In this study, we examine how to implement this controller in real-time using only a limited number of actuators. To do so, we formulate a convex quadratic programming problem that tunes the feedback gains of the controller in real-time such that it becomes realizable by the actuators. We evaluated the controller's performance through experiments on a physical soft robot capable of planar motions and show that the actual controller implemented by the finite-dimensional actuators still preserves the stabilizing property of the desired infinite-dimensional controller.
This research fills the gap between the infinite-dimensional control design and finite-dimensional actuation in practice and suggests a promising direction for exploring PDE-based control design for soft robots.


    

  
\end{abstract}

\section{Introduction}
Soft robots offer the ability to interact with the world in a more life-like manner. However, due to their inherent flexibility, soft robots are considered to have infinite degrees of freedom. Thus, modeling and control of soft robots pose a new set of challenges to robotics research. The creation of programmable soft bodies employing materials that incorporate sensors, actuators, and computing is the primary problem for developing soft machines that live up to their full potential\cite{survey}.

The most common strategy to control soft robots is using Piecewise Constant Curvature (PCC) models \cite{PCC}. This method assumes that a soft robotic arm is made up of a limited number of curved segments.  The major limitation of PCC models is that their accuracy suffers in instances of large deformation and external loading \cite{janabi2021cosserat}. Another approach is using finite element methods (FEM) to model and control soft robots \cite{FEM_overview}. FEM are numerical methods that approximate Partial Differential Equations through interactions between many local nodes. FEM  have the ability to model complex domains and have been shown to work in hardware \cite{FEM}, but these methods are extremely computationally expensive and it is quite difficult to implement in applications that require real-time responses.  Additionally, control design using FEM must rely on additional approximations such as quasistatic assumptions, linearization, and model reduction\cite{issues_FEM}, which can result in additional errors in the approximation.

The models developed from continuum mechanics are more precise, particularly the Cosserat rod theory for soft robots that resemble slender rods. Using a set of nonlinear partial differential equations (PDE), the Cosserat rod theory describes the time evolution of the infinite-dimensional kinematic variables of a deformable rod subject to external forces and moments. The aforementioned PCC and FEM models can be thought of as finite-dimensional approximations of the Cosserat PDE models\cite{discrete}.

Cosserat-rod PDEs have been shown to be more accurate by a series of experiments and widely employed as the basis of soft robot simulators \cite{simo1988dynamics}. However, the existing control design based on Cosserat-rod PDEs has mainly relied on discretization to obtain finite-dimensional ODEs due to the lack of efficient control theory for nonlinear PDEs. PDE-based nonlinear control can avoid modeling uncertainties due to discretization and yield more interpretable and computationally efficient controllers.

The vast majority of control methods for soft robots utilize open-loop control\cite{Open_Loop}. Open loop control methods are effective to demonstrate the feasibility of soft robots, but they do not allow the system to correct course if the real system deviates from the model. Closed-loop feedback control of soft robots with stability guarantees is still a challenging problem.

Recently, we presented an infinite-dimensional feedback controller for the Cosserat rod model and proved its stability \cite{tongjia}. In this work, we address the problem of how to implement such an infinite-dimensional controller in practice using specific actuators in real-time, which fills the gap between theory and experiment.
Specifically, we incorporate fabric series Pneumatic Artificial Muscle (fabric sPAM) actuators into the Cosserat-rod model and treat their air pressure as the actual inputs.
The feedback system states are estimated via computer vision.
The difficulty of implementing an infinite-dimensional feedback controller is that such a controller naturally lies in an infinite-dimensional functional space but the functional space that is implementable by a finite number of actuators is always finite-dimensional. 
To address this challenge, we formulate a convex quadratic programming problem to restrict the desired controller onto the actuators' implementable space and solve it to simultaneously determine the feedback gain functions and the actuator pressure. 
In this way, we guarantee that the actual feedback controller implemented by the actuators still preserves the stabilizing property of the desired infinite-dimensional controller.
Additional testing was conducted to maximize the convergence rate of the controller without producing an overshoot.


\par The paper is organized as follows. A planar version of the Cosserat Rod Theory is introduced in Section II. In Section III, the models for the actuators and gravity are introduced. Additionally, the PDE Controller is discussed as well as the formulation of an optimization problem to find the ideal pressure. Next, in Section IV, the physical testbed is discussed including the fabric sPAM actuators and the Computer Vision system. Then, in Section V, our results are discussed. Finally, in Section VI, we summarize our contribution and discuss future directions for this research.

\section{Modeling}
In this section, we introduce the PDE model for soft robots based on the Cosserat rod theory and incorporate fabric Series Pneumatic Artificial Muscle (fabric sPAM) actuators into the Cosserat-rod model for control purposes.

Cosserat-rod models are geometrically exact models that describe the dynamic response of long and thin deformable rods undergoing external forces and moments \cite{simo1988dynamics} and have been widely used to model soft manipulators \cite{rucker2011statics, renda2014dynamic, zheng2022pde}.
A Cosserat rod is idealized as a spatial curve consisting of four types of strains: bending, torsion, shear, and extension \cite{simo1988dynamics}.
For slender robots, the linear strains (shear and extension) have a negligible effect compared with the angular strains (bending and torsion) \cite{till2019real}.
Therefore, we can neglect these linear strain modes and obtain a reduced Cosserat-rod model which is also known as the Kirchhoff-rod model \cite{antman2005nonlinear}.

The modeling strategy and control algorithms described in this work are applicable to both the three-dimensional (3D) and two-dimensional (2D) systems, however, we will focus on the planar 2D case to simplify the notations and to explicitly show that the infinite-dimensional controllers can be approximated by a finite number of actuators in real-time. 

\subsection{Cosserat Rod Model for Planar Case}


Let $\{\Hat{y},\Hat{z}\}$ define a fixed orthonormal basis in the two-dimensional world frame. When the soft robot is in its unactuated state the length of the rod is defined as $L_0$ and the backbone lies along the z-axis. Because we are modeling a continuous backbone, all states depend on the independent variables of time $t\in\mathbb{R}$  and the arc-length of the center line $s\in[0,L_0]$. In this model, partial derivatives with respect to $t$ and $s$ are denoted by $\partial_t$ and $\partial_s$. The state of the rod can be described through the following vector:

$$q(s,t)\equiv\begin{bmatrix}y(s,t)\\ z(s,t)\\ \theta(s,t) \end{bmatrix}$$
where $r=(y,z)\in\mathbb{R}^2$ denotes the position vector of the centerline. Additionally, the angle $\theta\in R$ defines a local frame at each cross-section spanned by the orthonormal pair $\{j,k\}$, where $j=\cos\theta\Hat{y}-\sin\theta\Hat{z}$ and $k=\sin\theta\Hat{y}+\cos\theta\Hat{z}$. The vector $j$ is normal to each cross-section and is always tangent to the centerline of the robot.

\begin{figure}[!ht]
				        \centering
			        	\includegraphics[width=6cm]{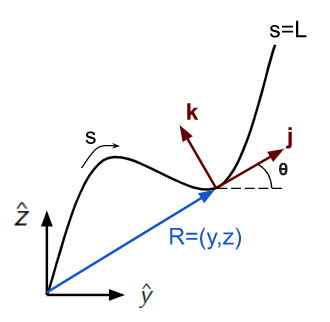}
				        \caption{\textbf{Reference frames used in Cosserat Rod Theory.} The global frame $\{y\}$ and the local frame described by}
				        \label{fig:vision}
				        \end{figure}

\par The forward kinematics of the system can be written as a function of $w_x$, the angular velocity about the $\Hat{x}$ direction:

$$\partial_tq=\begin{bmatrix} -z*w_x \\ x*w_x \\ w_x \end{bmatrix}$$

Here, $w_x$ is the angular velocity about $\Hat{x}$, which points out of the plane and completes the right handed coordinate system. The time rate of change of the angular velocity can be written as:

$$\partial_tw_x=\frac{1}{\rho J_x}(EJ_x\partial_s u+l_c(s)+l_g(s)).$$
where $u$ is the angular strain and is defined as $u=\partial_s\theta$. Additionally, $\rho$ is the density of the backbone, $J_x$ is the polar moment of inertia about the $\Hat{x}$ axis, $E$ is the Young's Modulus of the backbone, $l_c$ is the total moment generated by the pneumatic actuators, and $l_g$ is the moment due to gravity (the actual values of each parameter are given in Table 1). The  actuator and gravity model will be given in Section III.


When we combine all this information into one expression, the 1D manifestation of Cosserat Rod theory can be described in a single wave equation:
\begin{equation}\label{eq:wave equation}
    \partial_{tt}\theta=\frac{1}{\rho J_x}(EJ_x\partial_{ss} \theta+l_c(s)+l_g(s)).
\end{equation}

\subsection{Actuator Modeling}

In order to determine how much moment each chamber can apply to the system, we first model the amount of force each chamber can apply. We used the previous model that modeled the fabric sPAM actuators as ideal McKibben actuators \cite{FPAM}.
Although the system is not a perfect cylinder at the extremes of the chambers and near the O-rings, we neglect boundary effects and assume that the actuators have a cylindrical shape when inflated (Fig. 2). According to a previous study \cite{FPAM} which utilizes actuators made from the same material, the force that can be applied by a fabric sPAM actuator is modeled as follows: 
		        
$$F_{ideal}(\varepsilon)=(\pi r_0^2)P[a(1-\varepsilon)^2-b],\ 0\leq\varepsilon\leq\varepsilon_{max}$$        
where $a=3/\tan^2(\alpha_0)$ and $b=1/\sin^2(\alpha_0)$. Additionally, $P$ refers to the internal pressure, $\varepsilon$ refers to the contraction ratio, and $r_0$ and $\alpha_0$ refer to the initial depressurized radius and braid angle. The force and also the moment that the actuators can apply depend on the contraction of the actuators. 


\begin{figure}[!ht]
				        \centering
			        	\includegraphics[width=8cm]{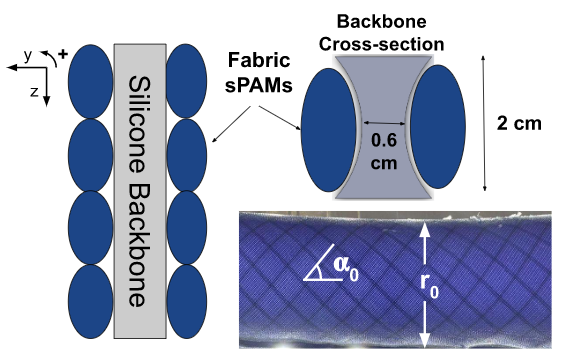}
				        \caption{{\bf Symbolic representation of soft manipulator} and visualization of the braid angle $\alpha_0$ and initial radius $r_0$.}
				        \label{fig:vision}
				        \end{figure}

When the fabric sPAM actuators are actuated, they can generate force in the negative z direction.  For both fabric sPAM actuators, the force acts along a line a fixed distance away from the center of the backbone. The distance from the center of the backbone to the center of the bladder is physically constrained to be d = 1.8 cm. Thus, when pressure is applied to an actuator, it contracts (shortens in length) and exerts torque in the same direction as the actuator.

$$l_c(s)=F_{ideal}(\varepsilon)d=(\pi r_0^2)P[a(1-\varepsilon)^2-b]d,\ 0\leq\varepsilon\leq\varepsilon_{max}$$

\subsection{Modeling Gravity}


In order to model the torque due to gravity, we must define a couple of terms. Torques must be defined around a reference point, so we define $s$ to be the reference point of interest at an arbitrary point along the length of the manipulator.

We also define a lever arm term for torque $r(s) \in  [-s,L-s]$. This measures the perpendicular distance between the force's line of action and the point of interest. Because we assume that the force of gravity always points straight downward (in the +z direction), this is just the horizontal distance between the reference point and the point at which the force occurs. 

The torque due to gravity is comprised of two components: the torque caused by the continuous soft backbone and the torque caused by the reaction force of the pin that holds the backbone up.

The torque due to the backbone can be calculated as follows:

$$l_{g,back}(s)=\int_{0}^{L} \rho A_c g*r(\sigma-s) \,d\sigma$$

where $\rho$ is the density of the back, $A_c$ is the cross-sectional area of the backbone, and g is the gravitational constant which is assumed to be $g=9.81 [m/s^2]$. In practice, this integral is approximated numerically through the known position of points along the length of the manipulator. Similarly, the torque due to the pin for the point of interest is computed as:

$$l_{g,pin}(s)=-\rho A_c L*r(-s)$$

Combining these terms, we obtain a complete expression for the torque due to gravity.

$$l_g(s)=l_{g,back}(s)+l_{g,pin}(s)$$
$$=\int_{0}^{L} \rho A_c g*r(\sigma-s) \,d\sigma-\rho A_c L*r(-s)$$

\section{Controller Design}
In this section, we review an infinite-dimensional controller from our previous work \cite{tongjia} and show how to tune some parameters of this infinite-dimensional controller such that it becomes realizable by only finitely many actuators.

Since shear and elongation deformations are ignored in the Kirchhoff case, the position of the soft robot can be uniquely determined by the angle $\theta$.
Thus, it suffices to consider the control problem for \eqref{eq:wave equation}.
After substituting the actuator and gravity models, we obtain the following complete control system:
\begin{align}\label{eq:complete system}
\begin{split}
    \partial_t\theta & =w_x \\
    \partial_tw_x & =l
\end{split}
\end{align}
where 
\begin{align}\label{eq:l*}
    l=\frac{1}{\rho J_x}(EJ_x\partial_{ss} \theta+(\pi r_0^2)P[a(1-\varepsilon)^2-b]d+l_g),
\end{align}
and the pressure $P$ is the actual input.
Assume the objective is to track the desired configuration trajectory given by
\begin{align*}
    \partial_t\theta_*=w_{*,x}.
\end{align*}
The position and velocity error terms are defined as:
$$e_\theta=\sin(\theta-\theta_*),\quad e_{w_x}=w_x-w_{*,x}.$$
Note that $e_\theta\to0$ implies $\theta\to\theta_*$.
In \cite{tongjia}, we presented an infinite-dimensional geometric controller for rotational control of 3D Cosserat rod models, which asymptotically drives the soft robot toward the desired configuration trajectory. In the planer case, the geometric controller is reduced to
\begin{equation}\label{eq:PD controller}
    l_*=\partial_{tt}\theta_*-k_\theta e_\theta-k_{w_x}e_{w_x},
\end{equation}
where $k_\theta(s),k_{w_x}(s,t)\in\mathbb{R}$ are feedback gains.
We have the following convergence result for this controller.
\begin{theorem}\label{thm:stability}
\cite{tongjia} Consider the soft robot system \eqref{eq:complete system}.
If there exist positive functions $k_\theta(s),k_{w_x}(s,t)$ such that $l\equiv l_*$ for all $t$, then $\big(e_\theta(s,t),e_{w_x}(s,t)\big)\to0$ for all $s$ exponentially.
\end{theorem}

This theorem implies that if we can design the pressure $P$ such that the actual input \eqref{eq:l*} has the form \eqref{eq:PD controller}, then we can guarantee the convergence of the configuration tracking objective.
The challenge is that the desired controller \eqref{eq:PD controller} is infinite-dimensional while the actual controller \eqref{eq:l*} always lies in a finite-dimensional functional space as $P$ is finite-dimensional.
If we prescribe positive values for $k_\theta,k_{w_x}$, it is almost impossible to find the correct air pressure $P$ such that \eqref{eq:l*} holds.
Nevertheless, we point out that $k_{w_x}$ is allowed to be a function of $(s,t)$.
Our key idea is to use this flexibility and tune the value of $k_{w_x}$ at every $t$ such that the corresponding desired controller \eqref{eq:PD controller} becomes realizable by the actual controller \eqref{eq:l*}.
This motivates us to formulate the following (convex) quadratic programming problem at every $t$:
\begin{align}\label{eq:programming}
\begin{split}
    \min_{P,k_{w_x}(s)} \quad & \|l-l_*\|_{L^2}^2 \\
    \text{s.t.} \quad & |P| \leq P_{max} \\
    & k_{w_x}(s)\geq\Bar{k},\quad \forall s,
\end{split}    
\end{align}
where $P_{max}$ is the maximum pressure and $\Bar{k}>0$ is a small constant.

Intuitively, this programming problem enforces the actual and desired controllers \eqref{eq:l*} and \eqref{eq:PD controller} to be equal by simultaneously determining the values of the actuator pressure $P$ and the feedback gain $k_{w_x}$ in real-time, subject to the constraints the $P$ is bounded and $k_{w_x}(s)$ is positive (for guaranteeing stability).

\begin{remark}
A few comments are in order.
First, the programming problem \eqref{eq:programming} is infinite-dimensional but can be numerically solved by discretization or parametrization.
Since it is a convex quadratic programming, there exist efficient commercial solvers that can solve it in real-time.
Second, whether a solution exists depends on whether the desired configuration is reachable given the actuator profile.
Our hypothesis is that a solution always exists if the desired static configuration is assignable.
This problem is under study.
Even if the desired configuration is not reachable, \eqref{eq:programming} still tries to find inputs that drive the soft robot as close as possible to the desired configuration.
Third, one may allow $k_\theta$ to also be a decision variable in \eqref{eq:programming} which will increase the possibility of finding a solution, but the stability results in Theorem \ref{thm:stability} no longer hold.
Nevertheless, convergence is still observed in experiments as long as the motion is sufficiently slow.
Finally, all results in this section can be easily extended to the 3D case by changing \eqref{eq:PD controller} to its 3D version given in \cite{tongjia}.
\end{remark}


\begin{table}[b!]
\caption{Physical parameters of the backbone of the soft manipulator.}
\centering
\begin{tabular}{ |p{3cm}|p{2cm}|p{2cm}|  }
\hline
Property & Symbol & Value \\
\hline
Density & $\rho$ & $1070\ \frac{kg}{m^3}$ \\
Young's Modulus & $E$ & $90\ kPA$ \\
Cross-sectional Area & $A_c$ & $1.68\cdot10^{-4}\ m^2$ \\
Moment of Area & $J_x$ & $0.12\cdot10^{-8}\ m^4$ \\
Length &  $L$ & $0.3\ m$ \\
\hline
\end{tabular}
\label{table:1}
\end{table}

\section{Experimental set-up}

\begin{figure}[!t]
				        \centering
			        	\includegraphics[width=8cm]{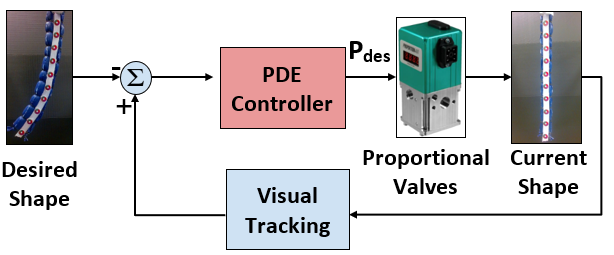}
				        \caption{{\bf Closed-loop shape tracking block diagram}}
				        \label{fig:1}
				        \end{figure}

Our experimental setup includes the soft manipulator, a frame from which the soft manipulator hangs, a Logitech HD Pro Webcam C920, and a Pressure Control Subsystem that consists of an Arduino Uno and two proportional QB3 regulators. The QB3 is a closed loop pressure regulator made up of a mechanical regulator mounted to two solenoid valves, an internal pressure transducer, and electronic controls. By turning on the solenoid valves, which pressurize the mechanical regulator's pilot, the pressure is controlled. Both valves regulate the exhaust and the inlet respectively. The soft manipulator is comprised of a flexible silicone backbone and two fabric sPAMs. As the robot moves, the webcam captures an image running at 30 Hz, and the pressure is calculated as a function of the error between the current configuration and the desired configuration.\\

\textit{\textbf{Fabric sPAM Actuators}}: The robot consists of two fabric sPAM actuators.
Fabric sPAMs are well suited for soft robotics because of their durability, ease of construction, and fast response time\cite{FPAM}.  
The actuators were made of a single layer of woven, air-tight material. The material is airtight, silicone and urethane-impregnated, rip-stop nylon often used in tents and tarps.  This fabric is reinforced in the radial direction like Mckibben actuators. When the pressure is turned off, the muscle is in its lengthened state. As the pressure is applied, the system expands radially and contracts axially. This applies force inward axially \cite{FPAM}. 


\par These actuators are fabricated by cutting rip-stop nylon at a $45^\circ$ bias, enclosing the fabric into a $r=1.5 \ [cm]$ cylinder, and sealing that cylinder with a Silicone glue (Smooth-On SilPoxy) based on the fabrication steps described in \cite{FPAM}. To increase contraction we add O-rings along the length of the robot. In order to maximize contraction, testing was conducted to determine the maximum O-ring spacing that would not result in inactive regions \cite{SPAM}. At an O-ring spacing of 3 cm, bulging was observed in each sub-chamber which meant there were no inactive regions.  

The final system consisted of two fabric sPAM actuators (L = 30 cm, D = 3 cm) made of rip-stop nylon. 3mm O-Rings 
were added in 3 centimeter segments along the length of the robot. These fabric sPAM chambers are attached to a flexible backbone. The backbone is 30 centimeters in length, has a cross-sectional area of $1.68\cdot \ 10^{-4} m^2$, and was made of the Dragon Skin 10 Silicone produced by Smooth-On. This material was selected because its flexibility would not inhibit actuation while having extremely high tensile strength subject to deformation.

\begin{figure}[!t]
    \centering
    \includegraphics[width=0.5\textwidth]{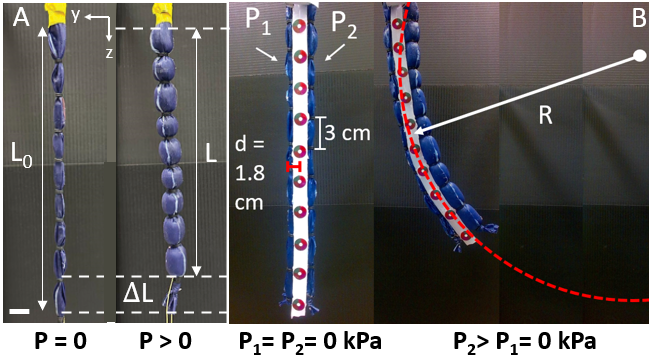}
    \caption{\textbf{Unactuated and actuated states of the fabric sPAMs.} \textbf{A.}  Unactuated (left, P = 0 kPa) and actuated (right, P = 30 kPa) states of a single fabric sPAM. \textbf{B.} Unactuated (left) and actuated (right) states of a two-segmented arm.  }
    \label{fig:my_label}
\end{figure}
\textbf{\textit{Pneumatic Control Board}}:
The pneumatic control board includes two Proportion Air QB3 proportional regulators with built-in pressure sensors.  This allows the regulator to control the closed loop pressure of each pressure channel to be the desired value. The desired pressure channel is controlled via a PWM signal from the Arduino. This PWM is converted to a true-analog signal via an analog conversion circuit. 


\textbf{\textit{Computer Vision State Observation}}: In order to calculate the posture of the soft robot, the position, and orientation of points of interest must be known. To do this, ten circular markers with four color quadrants were added along the length of the backbone. We first calculated the position of all ten points from a grey-scale version of the image using a circular Hough transform (Fig. \ref{fig:vision}). This was implemented using MATLAB's function  \textit{imfindcircles} from Image Processing Toolbox. Then we determined rotations of four lines connecting the central part of the marker and the centroids of all four color segments. Since the order of the color segments is known, we used these four angles to calculate the final marker's rotation angle. Using four color segments instead of one makes this approach more robust against noise. 

\begin{figure}[!t]
				        \centering
			        	\includegraphics[width=6cm]{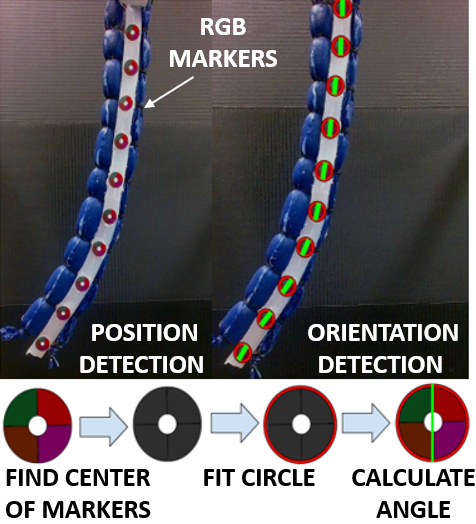}
				        \caption{{\bf Image processing example} (Left) An example image of the arm with markers. (right) Automatic detection of markers (red circles) and markers' orientation (green lines inside red circles).}
				        \label{fig:vision}
				        \end{figure}

\subsection{Estimating Contraction}
The force of the pneumatic artificial muscles is dependent on the contraction which can be calculated from the curvature of the backbone. We can derive the arc length of the backbone, $L=r\theta$, and the arc length of an actuator, $\quad L(1-\varepsilon)=(r-d)\theta$, as a function of the radius of curvature, $r$, and  the contraction ratio $\varepsilon=\Delta L/L$ (a unitless representation of the shrinkage of the chamber) (Fig. 4) . Since the backbone and chamber 2 are physically connected, they must share the same angle. This means that we can solve for $\theta$ in each equation and eliminate it from the system:
                    $$\frac{L}{r}=\frac{L(1-\varepsilon)}{r-d}\rightarrow r-d=r(1-\varepsilon)$$
                    $$\boxed{r=\frac{d}{\varepsilon}}\text{ or }\boxed{\varepsilon=\frac{d}{r}}$$
                    
Using this expression, we can apply known pressures to the system and find the radius of curvature of the corresponding section of the backbone. The radius of curvature is determined by using the Pratt Method proposed in 1987 to find the circle that best fits the curvature of the backbone \cite{pratt_1987}.

\subsection{Experimental Results}


The proposed controller is effectively an infinite-dimensional PD Controller. However, a full loop from image acquisition, to data processing, to setting pressure values takes about 0.5 seconds to complete. This means that any information about the angular velocity would not be applicable once the changes can be reflected in the robot. Thus, in practice, the dependence on the angular velocity of each point was eliminated making  the system an infinite dimensional proportional controller.

Although the controller will theoretically converge for any positive values of $k_\theta$, we want to maximize this value so the controller converges within a reasonable amount of time. To do this, We identified of the $k_\theta$ terms in optimization problem that would minimize the convergence time without excessive overshoot.

In general, the convergence time decreases as the lower bound increases. However, as the lower bound approaches $10^6$, overshoot starts to become a significant issue. In two of the five trials conducted, the soft manipulator took over two minutes to converge to the desired shape. In the instances that the system did converge with this lower bound, the soft manipulator converges to the desired location the fastest out of any trial. However, if the system got stuck continuously overshooting the target, this was considered a failure for the controller. The decision was made to trade precision for convergence speed and a lower bound of $10^5$ was selected.

\begin{figure}[!t]
\centering
\includegraphics[width=8 cm]{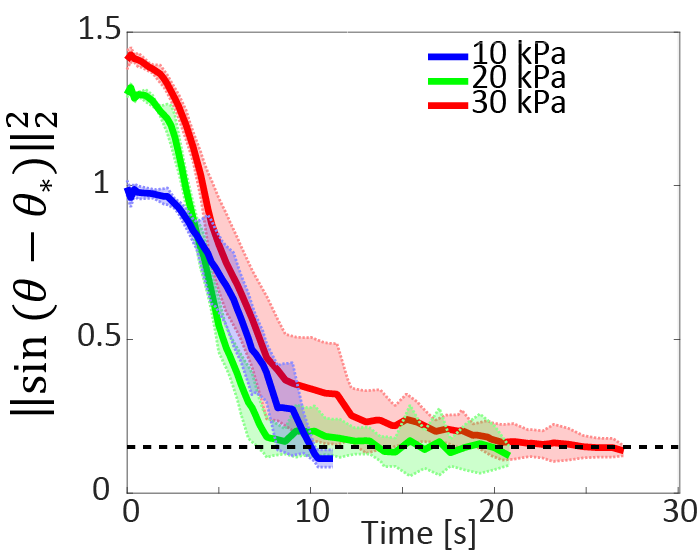}
\caption{{\bf{Posture Convergence}} The mean response for the norm of the angular error vector over time. Five trials were conducted at each pressure (blue = 10 kPa, green = 20 kPa, red = 30 kPa) and the error band show one standard deviation from the mean as a function of time. The dashed line shows the stopping criteria. }
\label{fig}
\end{figure}

Several tests were conducted to demonstrate the controller's ability to converge to a desired configuration. 
The plot in Fig. 6 shows the error of the system decaying
over time so that the norm of the error vector is within 0.15 of the
desired configuration. This 0.15 stopping criterion was determined by the reliability of the computer vision system. Theoretically, the system will remain fixed within a plane. However, small imperfections cause the manipulator to twist slightly, especially at higher pressures. Because of error due to twisting, the stopping criterion was implemented.


\begin{figure}
    \centering
    \includegraphics[width=8cm]{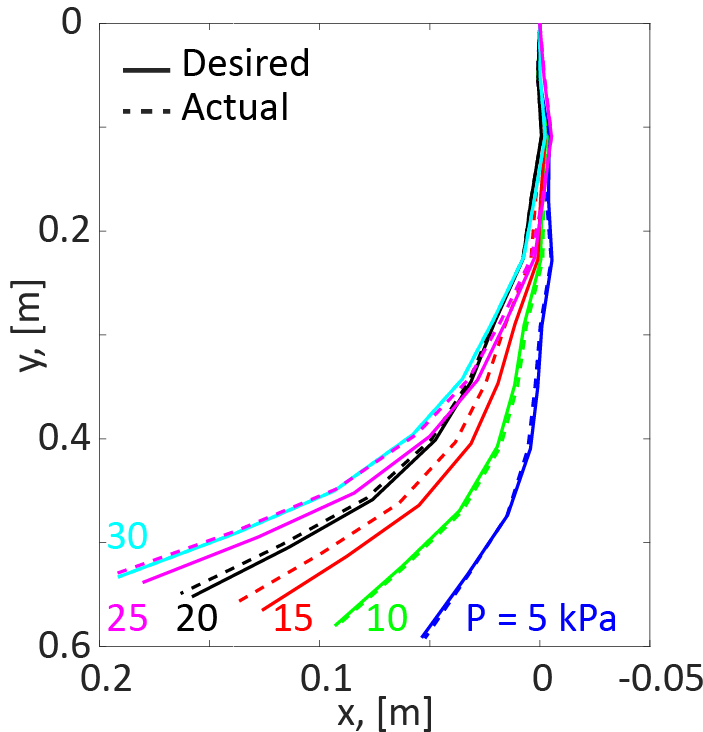}
    \caption{{\bf Shape Replication} Given a orientation associated with a known pressure, the PDE Controller is able to replicate the shape of robot for pressure from 5-30 kPa }
    \label{fig:my_label}
\end{figure}

Next, we observed how the controller performed across a range of pressure values as shown in Fig. 7. For all the pressure ranges, the found configuration corresponds to the desired shape quite well. For low pressure values, the actual configuration is nearly identical to the desired configuration. The shape deviates more for pressure above 20 kPa. We attribute this to two factors. The first is the twisting behavior that causes errors in computer vision estimation at higher pressure. Additionally, the majority of contraction occurs between 0-20 kPa, which means that above this range increase beyond 20, the pressure that corresponds to the stopping criterion will deviate more from the known pressure. However, the estimated shapes on the high end of the pressure are still reasonable estimates of the shape.

Overall, these results show promise for the feasibility of partial differential control in hardware. With the development of more sophisticated computer vision methods and fine tuned models, the hope is to reliably demonstrate the ability to control soft robots of any kind with this technique.


\section{Conclusion}
In this work, we demonstrated the efficacy of partial differential equation feedback control using a finite number of actuators in real time for the shape control of a planar soft robot. In order to do this, we presented a simplified version of Cosserat Rod Theory, manufactured a two-chamber pneumatic soft robot, and developed a computer vision algorithm to observe the state of the system. In future work, we plan to demonstrate that this controller works for other pneumatic actuators and extend the results to three dimensions.

\section{Awknowledgements}
We would like to thank Adam Czajka of the University of Notre Dame for his help with computer vision topics.

\bibliographystyle{IEEEtran}

\end{document}